\documentclass[journal]{IEEEtran}
\usepackage{amsmath,amsfonts}
\usepackage{url}
\usepackage[hidelinks,hypertexnames=false]{hyperref}
\usepackage{cite}
\usepackage{booktabs}
\usepackage{multirow}
\usepackage{makecell}
\usepackage{enumitem}
\usepackage{ulem}
\usepackage{color}
\usepackage[table,xcdraw]{xcolor}
\usepackage{scalerel}
\usepackage{tikz}
\usetikzlibrary{svg.path}

\definecolor{orcidlogocol}{HTML}{A6CE39}
\tikzset{
  orcidlogo/.pic={
    \fill[orcidlogocol] svg{M256,128c0,70.7-57.3,128-128,128C57.3,256,0,198.7,0,128C0,57.3,57.3,0,128,0C198.7,0,256,57.3,256,128z};
    \fill[white] svg{M86.3,186.2H70.9V79.1h15.4v48.4V186.2z}
                 svg{M108.9,79.1h41.6c39.6,0,57,28.3,57,53.6c0,27.5-21.5,53.6-56.8,53.6h-41.8V79.1z M124.3,172.4h24.5c34.9,0,42.9-26.5,42.9-39.7c0-21.5-13.7-39.7-43.7-39.7h-23.7V172.4z}
                 svg{M88.7,56.8c0,5.5-4.5,10.1-10.1,10.1c-5.6,0-10.1-4.6-10.1-10.1c0-5.6,4.5-10.1,10.1-10.1C84.2,46.7,88.7,51.3,88.7,56.8z};
  }
}
\newcommand\orcidicon[1]{\href{https://orcid.org/#1}{\mbox{\scalerel*{
\begin{tikzpicture}[yscale=-1,transform shape]
\pic{orcidlogo};
\end{tikzpicture}
}{|}}}}
\newcolumntype{L}[1]{>{\raggedright\arraybackslash}p{#1}}
\newcolumntype{C}[1]{>{\centering\arraybackslash}p{#1}}
\newcolumntype{R}[1]{>{\raggedleft\arraybackslash}p{#1}}

\definecolor{TYS}{rgb}{0.6, 0.8, 0.2}
\definecolor{DOcolor}{rgb}{1,0.45,0.0}
\definecolor{NAVYcolor}{rgb}{0.05,0,0.5}
\newcommand{\settablefont}{\fontsize{6.5}{11.8}\selectfont}

\begin{document}
\title{RoadFormer: Duplex Transformer for RGB-Normal\\Semantic Road Scene Parsing}

\normalem
\author{Jiahang Li$^{\orcidicon{0009-0005-8379-249X}\,}$,~\IEEEmembership{Graduate Student Member,~IEEE}, Yikang Zhang$^{\orcidicon{0009-0003-8840-392X}\,}$,~\IEEEmembership{Graduate Student Member,~IEEE},
\\Peng Yun$^{\orcidicon{0000-0002-8163-267X}\,}$, Guangliang Zhou$^{\orcidicon{0000-0001-7845-7068}\,}$, Qijun Chen$^{\orcidicon{0000-0001-5644-1188}\,}$~\IEEEmembership{Senior Member,~IEEE}, Rui Fan$^{\orcidicon{0000-0003-2593-6596}\,}$,~\IEEEmembership{Senior Member,~IEEE}
\thanks{This research was supported by the National Science and Technology Major Project under Grant 2020AAA0108101, the National Natural Science Foundation of China under Grant 62233013, the Science and Technology Commission of Shanghai Municipal under Grant 22511104500, the Fundamental Research Funds for the Central Universities, and Xiaomi Young Talents Program. (\emph{Corresponding author: Rui Fan})}
\thanks{
Jiahang Li, Yikang Zhang, Guangliang Zhou, Qijun Chen, and Rui Fan are with the College of Electronics \& Information Engineering, Shanghai Research Institute for Intelligent Autonomous Systems, the State Key Laboratory of Intelligent Autonomous Systems, and Frontiers Science Center for Intelligent Autonomous Systems, Tongji University, Shanghai 201804, P. R. China (e-mails: \{lijiahang617, yikangzhang, tj\_zgl, qjchen\}@tongji.edu.cn; {rui.fan@ieee.org}).
}
\thanks{Peng Yun is with the Department of Computer Science and Engineering, The Hong Kong University of Science and Technology, Hong Kong. (e-mail: { pyun@connect.ust.hk}).}
}

\markboth{IEEE Transactions on Intelligent Vehicles}{}

\maketitle

\begin{abstract}
The recent advancements in deep convolutional neural networks have shown significant promise in the domain of road scene parsing. Nevertheless, the existing works focus primarily on freespace detection, with little attention given to hazardous road defects that could compromise both driving safety and comfort. In this article, we introduce RoadFormer, a novel Transformer-based data-fusion network developed for road scene parsing. RoadFormer utilizes a duplex encoder architecture to extract heterogeneous features from both RGB images and surface normal information. The encoded features are subsequently fed into a novel heterogeneous feature synergy block for effective feature fusion and recalibration. The pixel decoder then learns multi-scale long-range dependencies from the fused and recalibrated heterogeneous features, which are subsequently processed by a Transformer decoder to produce the final semantic prediction. Additionally, we release SYN-UDTIRI, the first large-scale road scene parsing dataset that contains over 10,407 RGB images, dense depth images, and the corresponding pixel-level annotations for both freespace and road defects of different shapes and sizes. Extensive experimental evaluations conducted on our SYN-UDTIRI dataset, as well as on three public datasets, including KITTI road, CityScapes, and ORFD, demonstrate that RoadFormer outperforms all other state-of-the-art networks for road scene parsing. Specifically, RoadFormer ranks first on the KITTI road benchmark. Our source code, created dataset, and demo video are publicly available at \url{mias.group/RoadFormer}. 
\end{abstract}
\begin{IEEEkeywords}
Convolutional neural network, road scene parsing, freespace detection, semantic segmentation, driving safety and comfort, Transformer.
\end{IEEEkeywords}

\section{Introduction}
\label{sec.intro}
\IEEEPARstart{T}{he} advancements in machine intelligence have led to the extensive integration of autonomous driving technologies into various aspects of daily life and across multiple industries \cite{fan2020sne}. This integration spans a diverse range of products, including autonomous vehicles \cite{geiger2013vision}, mobile robots \cite{li2022towards}, and smart wheelchairs \cite{wang2021dynamic}. Recently, researchers in this field have shifted their focus towards enhancing both driving safety and comfort \cite{du2018velocity}. Road scene parsing, typically including pixel-level freespace and road defect detection, is of paramount importance in achieving these objectives \cite{barabino2019standing}.

Existing road scene parsing approaches predominantly fall under two categories: geometry-based and data-driven ones \cite{fan2020sne}. The algorithms in the former category typically leverage explicit geometric models to represent regions of interest (RoIs), and then proceed to optimize specific energy functions for accurate RoI extraction. For instance, the study presented in \cite{wedel2009b} employs a B-spline model to fit the road disparity map, which is subsequently projected onto a 2D v-disparity histogram for freespace detection. Additionally, the research presented in \cite{fan2019pothole} introduces a disparity map transformation algorithm designed specifically for the effective detection of road defects. 
Nonetheless, actual roads are often uneven, rendering such approaches occasionally infeasible \cite{ma2022computer}.

With the proliferation of deep learning techniques, convolutional neural networks (CNNs) have emerged as a potent tool for road scene parsing, often treating it as a binary or ternary semantic segmentation task \cite{fan2020sne,fan2021graph}. These methods have demonstrated significant performance gains over traditional geometry-based approaches. For example, the study detailed in \cite{lu2019monocular} employs an encoder-decoder CNN architecture to realize freespace detection by segmenting RGB images in the bird's eye view. Nonetheless, the results achieved by this approach fall short of satisfactory performance benchmarks. To address this limitation, ensuing research has explored the utilization of data-fusion networks with duplex encoder architectures as a feasible strategy to improve the road scene parsing accuracy. Specifically, \cite{hazirbas2017fusenet} extracts heterogeneous features from RGB-Depth data, and subsequently performs feature fusion through a basic element-wise summation operation. The fusion of disparate feature types brings a more comprehensive understanding of the given scenario, resulting in superior performance over earlier single-modal networks. Similar to \cite{hazirbas2017fusenet}, SNE-RoadSeg series \cite{fan2020sne, wang2021sne} performs RGB-Normal feature fusion through element-wise summation. By employing a duplex ResNet \cite{he2016deep} in conjunction with a strong densely-connected decoder, the SNE-RoadSeg series achieves state-of-the-art (SoTA) performance on the KITTI road benchmark \cite{fritsch2013new}. Nevertheless, a current bottleneck exists in the simplistic and indiscriminate fusion of heterogeneous features, often causing conflicting feature representations and erroneous detection results.

Within the domain of computer vision, Transformers have empirically demonstrated their potential to outperform CNNs, particularly when large-scale datasets are available for training  \cite{han2022survey, liu2021swin, xie2021segformer}. This advantage can be attributed to the self-attention mechanisms inherent to Transformers, which provide a notably more efficient strategy for global context modeling compared to conventional CNNs \cite{li2023uniformer}. Therefore, employing attention mechanisms to improve the fusion of heterogeneous features extracted by the duplex encoder is an area of research gaining popularity and deserving further attention. OFF-Net \cite{min2022orfd} is the first attempt to apply the Transformer architecture for road scene parsing. With abundant off-road training data, it outperforms CNN-based algorithms. Unfortunately, OFF-Net utilizes a lightweight CNN-based decoder instead of a Transformer-based one. We believe that the adoption of a Transformer-based decoder has the potential to elevate the upper limit of road scene parsing performance. We also observe its unsatisfactory performance on urban road scenes, especially when the data are limited. 

Therefore, in this article, we introduce RoadFormer, a novel duplex Transformer architecture designed for data-fusion semantic road scene parsing. Benefiting from its duplex encoder, RoadFormer can extract abstract and informative heterogeneous features from RGB-Normal data. Additionally, we introduce a novel Heterogeneous Feature Synergy Block (HFSB), which draws upon the self-attention mechanism to improve feature fusion and recalibration. Extensive experiments conducted on public freespace detection datasets demonstrate that RoadFormer achieves improved overall performance than existing SoTA networks. Specifically, \uline{RoadFormer ranks first on the KITTI road benchmark upon submission} \cite{fritsch2013new}. In contrast to existing studies and their utilized datasets in the domain of road scene parsing \cite{sun2020scalability, geiger2013vision, cordts2016cityscapes}, which predominantly characterize freespace as undamaged, the aspect of road defects remains notably under-explored. This oversight can be attributed, in part, to the sporadic nature of road defects, making the collection of comprehensive, large-scale datasets containing both RGB and depth images challenging. To this end, we create SYN-UDTIRI, a large-scale, multi-source synthetic dataset, specifically for the understanding of road scenes inclusive of defects. By publishing this dataset, we not only advance the scope of semantic parsing in road scenes but also open up new avenues for data-driven research, thereby demonstrating the capabilities of our proposed RoadFormer. 

In summary, the main contributions of this article include: 
\begin{itemize}
    \item RoadFormer, a Transformer-based data-fusion network is developed for semantic road scene parsing, where surface normal information is introduced to bring a more comprehensive understanding of road scenes, our RoadFormer achieves SoTA performance on the KITTI Road benchmark.
    \item A novel heterogeneous feature synergy block based on self-attention mechanism is designed to dynamically fuse RGB and normal features. Compared to other famous pure CNN-based attention modules, HFSB yields better results when paired with our RoadFormer.
    \item SYN-UDTIRI, a large-scale synthetic dataset is created for comprehensive road scene parsing. It contains 10,407 pairs of RGB and depth images, as well as pixel-level semantic annotations for freespace and road defects. We believe this dataset and benchmark could bridge the research gap for data-fusion road scene parsing, especially for road defect detection.
\end{itemize}

The remainder of this article is organized as follows: 
Sect. \ref{sec.related_works} reviews related works.
Sect. \ref{sec.methodology} details our proposed RoadFormer.
Sect. \ref{sec.experiments} presents the experimental results and compares our network with other SoTA single-modal and data-fusion models.
Finally, we conclude this article and discuss possible future work in Sect. \ref{sec.conclusion}.

\section{Related Work}
\label{sec.related_works}

\subsection{Single-Modal Networks}
\label{sec.single_modal}
Fully convolutional network (FCN) \cite{long2015fully} pioneered the utilization of CNNs for single-modal semantic segmentation. Although FCN significantly outperforms traditional methods based on hand-crafted features, its performance is constrained by the absence of multi-scale feature utilization \cite{ding2018context}. To address this limitation, DeepLabv3+ \cite{chen2018encoder} employs atrous convolutions with different dilation rates, thereby enhancing the network's capability to encode contextual features across multiple scales. The high-resolution network (HRNet) \cite{wang2020deep} employs an alternative strategy for multi-scale feature encoding. Instead of solely performing detrimental resizing of feature maps, HRNet integrates low-resolution branches in parallel with the high-resolution main branch to achieve multi-scale feature extraction.

Transformers have been applied to semantic segmentation tasks due to their superior global aggregation capabilities over CNNs \cite{strudel2021segmenter}. Segmentation Transformer (SETR) \cite{zheng2021rethinking} is the first Transformer-based general-purpose semantic segmentation network. Building on the success of Vision Transformer (ViT) \cite{dosovitskiy2020image}, it tokenizes images into patches and feeds them into Transformer blocks. These encoded features are then gradually restored through upsampling convolution to achieve pixel-level classification. SegFormer \cite{xie2021segformer} introduces a multi-scale Transformer encoder for semantic segmentation, which stacks Transformer blocks and inserts convolutional layers (for feature map downsampling) between them. Compared to SETR, SegFormer improves segmentation performance when dealing with objects of varying sizes. Additionally, MaskFormer \cite{cheng2021per} introduces a new paradigm for semantic segmentation. Rather than adhering to standard per-pixel classification methods, this architecture addresses the semantic segmentation task by decoding query features into a set of masks. In detail, MaskFormer employs a multi-scale Transformer decoder that outputs the mask for each class using refined queries in parallel, outperforming previous per-pixel classification approaches. This query-based prediction fashion is also adopted in our proposed method and has demonstrated improved performance over SoTA CNNs in terms of segmentation accuracy.

\subsection{Data-Fusion Networks}
\label{sec.data_fusion}
Data-fusion networks effectively leverage heterogeneous features extracted from RGB images and spatial geometric information to improve segmentation performance. FuseNet \cite{hazirbas2017fusenet} was the first attempt to incorporate depth information into semantic segmentation, using separate CNN encoders for RGB and depth images and fusing their features through element-wise summation. MFNet \cite{ha2017mfnet} strikes a balance between speed and accuracy in driving scene parsing through RGB-Thermal fusion. Similarly, RTFNet \cite{sun2019rtfnet} also uses RGB-Thermal data as inputs and designs a robust decoder that utilizes shortcuts to produce clear boundaries while retaining detailed features. Inspired by \cite{hazirbas2017fusenet}, SNE-RoadSeg \cite{fan2020sne} and SNE-RoadSeg+ \cite{wang2021sne} incorporate surface normal information into freespace detection. This series employs densely connected skip connections to enhance feature extraction in their decoder, thereby achieving SoTA performance compared to other approaches. Drawing on the success of single-modal Transformer models, OFF-Net \cite{min2022orfd} was the first attempt to apply Transformer architecture for data-fusion freespace detection. It utilizes a SegFormer \cite{xie2021segformer} encoder to generate RGB and surface normal features, outperforming SoTA CNNs in off-road freespace detection. Expanding upon these foundational prior arts, our RoadFormer also adopts the data-fusion paradigm but differentiates itself by employing a novel Transformer architecture for semantic road scene parsing. Additionally, RoadFormer utilizes a novel feature synergy block, leading to superior performance over all other data-fusion networks on four road scene parsing datasets.

\section{Methodology}
\label{sec.methodology}
\begin{figure*}[!ht]

\includegraphics[width=0.99\textwidth]{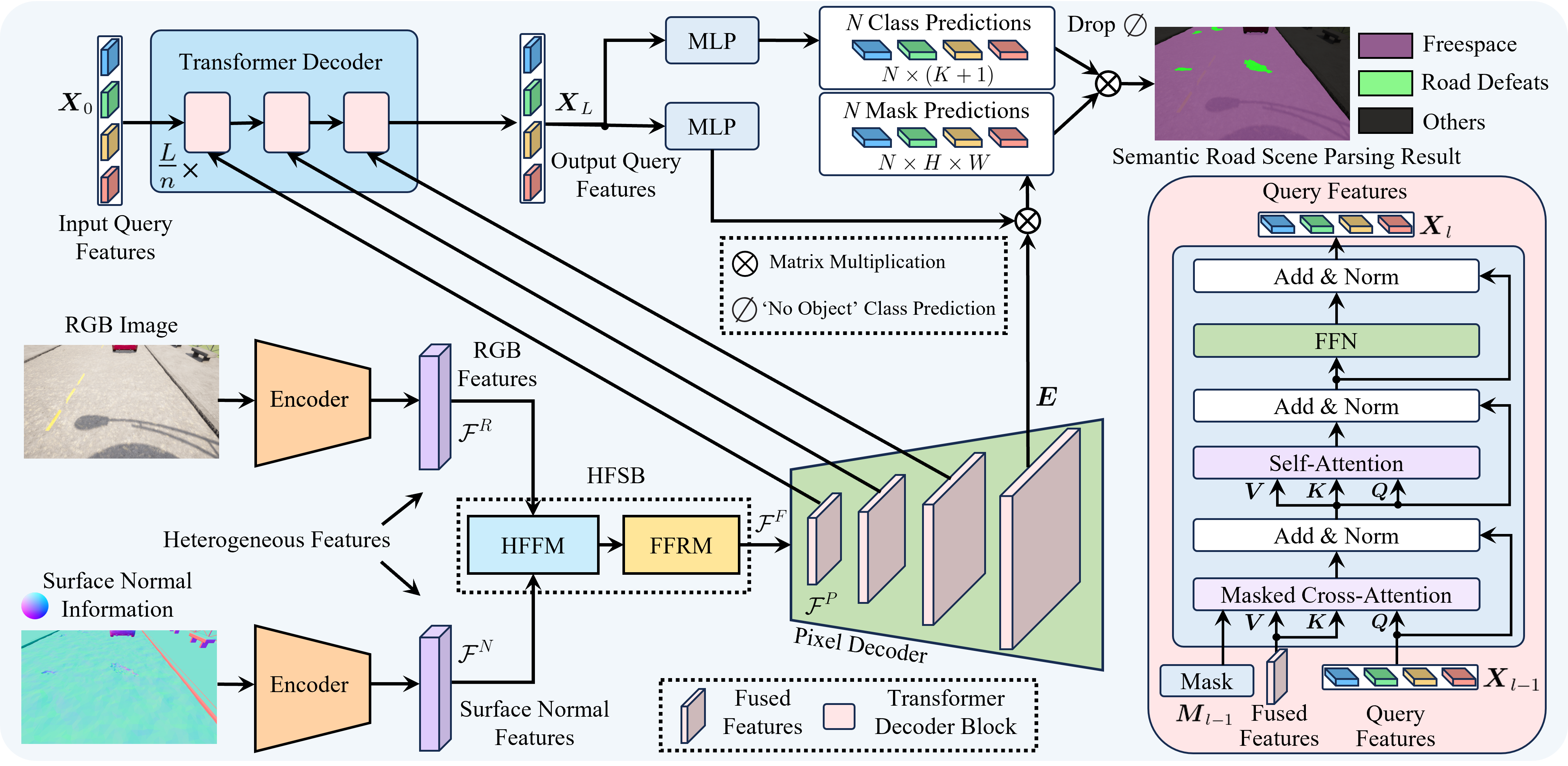}
\caption{An overview of our proposed RoadFormer architecture.}
\label{fig.structure}

\end{figure*}
This section details RoadFormer, a robust and powerful data-fusion Transformer architecture for road scene parsing. As depicted in Fig. \ref{fig.structure}, RoadFormer consists of:
\begin{enumerate}[label=(\arabic*)]
    \item a duplex encoder to learn heterogeneous features from RGB-Normal data;
    \item a feature synergy block to fuse and recalibrate the encoded heterogeneous features;
    \item a pixel decoder to learn long-range dependencies from recalibrated features;
    \item a Transformer decoder to achieve final semantic prediction by refining query features using outputs of the pixel decoder.
\end{enumerate}

\subsection{Duplex Encoder}
\label{sec.duplex_encoder}
In line with our previous works \cite{fan2020sne, wang2021sne, wang2021dynamic}, we employ a duplex encoder structure to extract multi-scale heterogeneous features. One encoder focuses on learning color and texture features $\mathcal{F}^{R}=\{ \boldsymbol{F}^R_1, \dots,\boldsymbol{F}^R_k \}$ from RGB images $\boldsymbol{I}^{R}\in\mathbb{R}^{H\times W \times 3}$, while the other encoder specializes in acquiring representations $\mathcal{F}^{N} = \{ \boldsymbol{F}^N_1, \dots, \boldsymbol{F}^N_k \}$ of planar characteristics from the surface normal information $\boldsymbol{I}^{N}\in\mathbb{R}^{H\times W \times 3}$, where $H$ and $W$ represent the height and width of the input image, respectively, $\boldsymbol{F}^{R,N}_i \in \mathbb{R}^{\frac{H}{S_i} \times \frac{W}{S_i} \times C_{i}}$ represents the $i$-th feature maps, and $C_{i}$ and $S_i=2^{i+1}$ ($i\in[1,k]\cap\mathbb{Z}$) denote the corresponding channel and stride numbers, respectively (usually $k = 4$). Our duplex encoder is compatible with both Transformer-based and CNN-based backbones. For this study, we utilize Swin Transformer \cite{liu2021swin} and ConvNeXt \cite{liu2022convnet} as our backbones, respectively. The performance comparison of these backbones is presented in Sect. \ref{sec.ablation_study}.

\subsection{Heterogeneous Feature Synergy Block}
\label{sec.hfsb}
\subsubsection{Heterogeneous Feature Fusion Module (HFFM)}
\begin{figure}
    \includegraphics[width=0.475\textwidth]{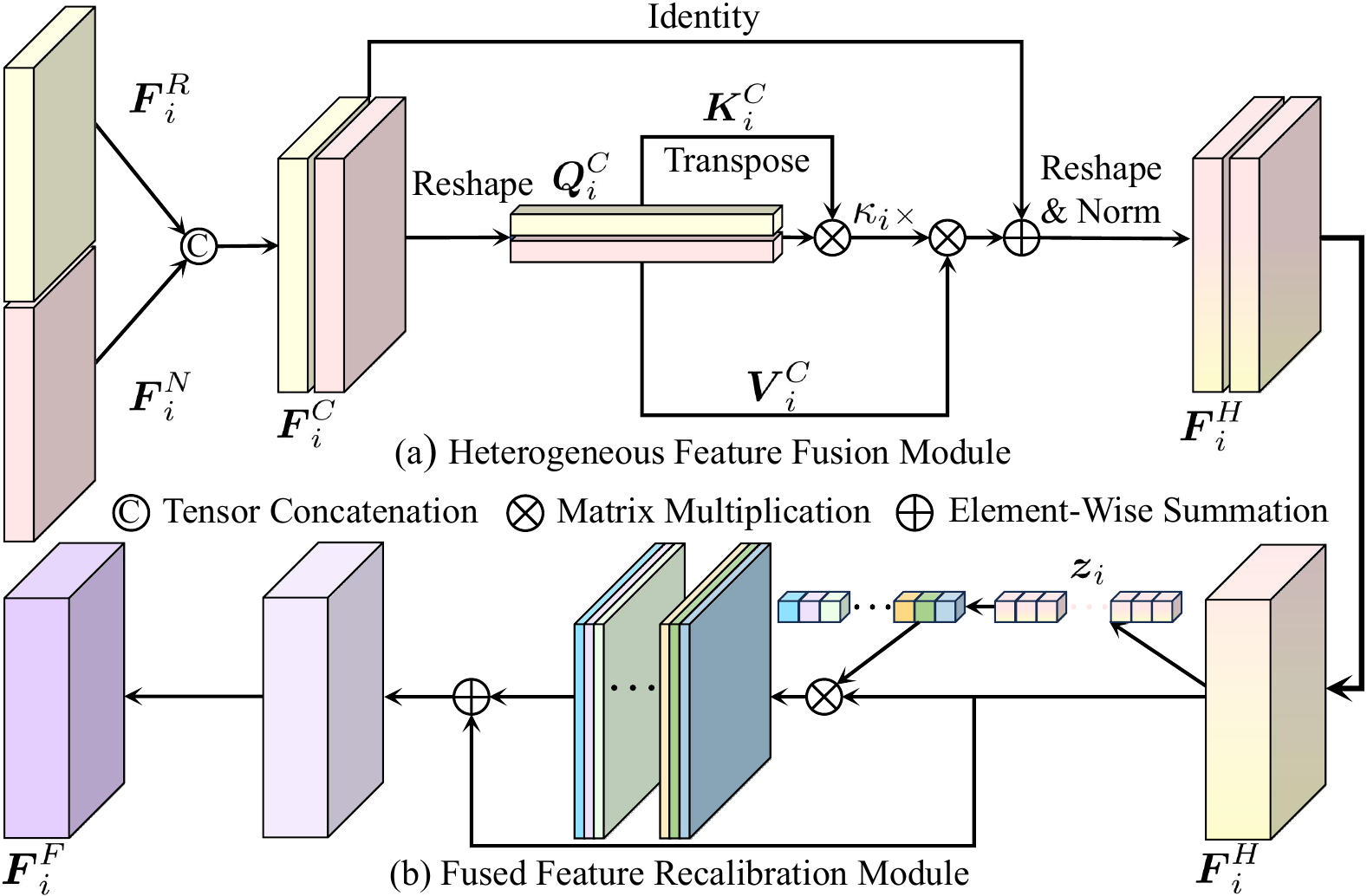}
        \caption{Heterogeneous feature synergy block.}
    \label{fig.two_modules}
\end{figure}

Conventional duplex networks typically fuse heterogeneous features using basic element-wise addition or feature concatenation operations \cite{wang2021dynamic}. Nevertheless, we contend that such a simplistic feature fusion strategy might not fully exploit the inherent potential of heterogeneous features. The recent surge in the popularity of Transformer architectures for multi-modal visual-linguistic tasks \cite{radford2021learning, brown2020language} can be attributed to their powerful attention mechanisms that enable effective joint representations across modalities \cite{xu2023multimodal}. Drawing inspiration from these prior arts, our objective is to utilize attention mechanisms to enhance the fusion of heterogeneous features extracted by the duplex encoder mentioned above. To this end, we introduce HFFM (see Fig. \ref{fig.two_modules} (a)), which builds upon the self-attention mechanisms of Transformers to achieve effective fusion of $\mathcal{F}^{R}$ and $\mathcal{F}^{N}$ through token interactions. Our HFFM can be formulated as follows:
\begin{equation}
     \boldsymbol{F}^H_i =  \mathrm{Reshape}\Big(\mathrm{Norm}\big(\mathrm{Softmax}
     ({\boldsymbol{Q}^C_i {\boldsymbol{K}^C_i}^\top}) \kappa_i\boldsymbol{V}^C_i +\boldsymbol{F}^C_i \big)\Big),
    \label{eq.self-attention2}
\end{equation} 
where $\boldsymbol{F}^R_i$ and $\boldsymbol{F}^N_i$ are concatenated and reshaped to form $\boldsymbol{F}^C_i \in \mathbb{R}^{2C_i \times \frac{H}{S_i}\frac{W}{S_i}}$, which is then identically mapped to query $\boldsymbol{Q}_i^C$, key $\boldsymbol{K}_i^C$, and value $\boldsymbol{V}_i^C$ embeddings, and $\boldsymbol{F}^H_i \in \mathbb{R}^{\frac{H}{S_i} \times \frac{W}{S_i} \times 2C_i}$ denotes the output of the HFFM. Additionally, following \cite{fu2019dual}, we introduce a learnable coefficient $\kappa_i$ to adaptively adjust the attention significance, enabling a more flexible fusion of heterogeneous features.

\subsubsection{Fused Feature Recalibration Module (FFRM)}
 \label{Sec.ffrm}
In conventional single-encoder architecture designs, multi-channel features do not all contribute positively to semantic predictions. In fact, some noisy or irrelevant feature maps may even degrade the model's performance. In this regard, the squeeze-and-excitation block (SEB) \cite{hu2018squeeze} was designed to model the inter-dependencies between the channels of convolutional features via a channel attention mechanism. This allows the network to focus on more informative features while downplaying the less important ones. A similar issue also occurs in our work: the heterogeneous features extracted by our duplex encoder may focus on different components in the scene, and the fusion of these features might potentially undermine the saliency of the original key features or even produce irrelevant features \cite{mai2021analyzing}. To this end, we develop FFRM (see Fig. \ref{fig.two_modules} (b)) based on SEB to recalibrate the fused heterogeneous features. In particular, we add a residual connection atop the SEB to enhance its training \cite{he2016deep} and introduce an additional point-wise convolution to realize the flexible computation of correlations between the recalibrated features \cite{chollet2017xception}. Our FFRM can be formulated as follows:
\begin{equation}
    \boldsymbol{{F}}^F_i 
    = {\underset{1\times 1}{\mathrm{Conv}}}
    \bigg(\boldsymbol{F}^H_i + 
    \Big(
    \boldsymbol{O}\ \mathrm{Sigmoid}
    \big(
    {\underset{1\times 1}{\mathrm{Conv}}}
    (\boldsymbol{z}_i)
    \big)
    \Big) 
    \odot \boldsymbol{F}^H_i
    \bigg),
\label{eq:dcwfr1}
\end{equation}
where $\boldsymbol{O}\in\mathbb{R}^{\frac{H}{S_i}\times \frac{W}{S_i} \times 1\times 1}$ represents a matrix of ones, $\odot$ denotes the Hadamard product operation, $\boldsymbol{z}_i=[z_{i,1},\dots,z_{i,2C_i}]\in\mathbb{R}^{1\times 1\times 2 C_i}$ stores the average pooling results of each feature map in $\boldsymbol{F}^H_{i}$, and 
\begin{equation}
    {z}_{i,j}= \frac{S_i^2}{H W} \sum_{h=1}^{\frac{H}{S_i}}\sum_{w=1}^{\frac{W}{S_i}}\boldsymbol{F}^H_{i}(h,w,j).
    \label{eq:dcwfr4}
\end{equation}
The performance comparison between our proposed HFFM, FFRM, and SE block is discussed in Sect. \ref{tab.modules_ablation}.

\subsection{Pixel Decoder}
\label{sec.pixel_decoder}
Following \cite{cheng2022masked}, we incorporate a pixel decoder to improve multi-scale feature modeling for {$\mathcal{F}^{F}=\{ \boldsymbol{F}^F_1, \dots,\boldsymbol{F}^F_k \}$} to generate $\mathcal{F}^{P}=\{ \boldsymbol{F}^P_1, \dots,\boldsymbol{F}^P_{n} \}$ ($n<k$, usually $n = 3$), where $\boldsymbol{F}^P_i\in\mathbb{R}^{\frac{H}{S_i}\times \frac{W}{S_i} \times C}$. It also serves the function of upsampling low-resolution features in {$\mathcal{F}^{F}$} to generate a high-resolution per-pixel embedding $\boldsymbol{E}$, which is then used by the Transformer decoder to guide the mask prediction. Given the recent success of multi-scale deformable attention Transformer \cite{zhu2020deformable, jain2023oneformer, cheng2022masked}, we adopt this architecture within our pixel decoder to generate $\mathcal{F}^{P}$. In contrast to the single-modal approach presented in \cite{cheng2022masked}, our pixel decoder processes the fused features $\mathcal{F}^{F}$ generated from both RGB images and surface normal information. Consequently, the feature map channel at each scale is doubled compared to \cite{cheng2022masked}. To reduce the increased computational complexity, we employ $1 \times 1$ convolutional layers to reduce the feature map channels before performing multi-scale deformable attention operations.

\subsection{Transformer Decoder}
\label{sec.transformer_decoder}

We utilize a Transformer-based decoder to recursively update the input query features $\boldsymbol{X}_0 \in \mathbb{R}^{N \times C}$ based on the multi-scale feature maps $\boldsymbol{F}^P_1$ to $\boldsymbol{F}^P_{n}$ extracted by the pixel decoder. The per-pixel embedding $\boldsymbol{E}$ is used to guide mask predictions. 
Specifically, $N$ learnable feature vectors with $C$ channels are initialized as the input query features ${\boldsymbol{X}_{0}}$, which are fed into the subsequent Transformer decoder layers.
One Transformer decoder layer consists of a sequence of operations: (1) the query $\boldsymbol{Q}^D_l = f_Q({\boldsymbol{X}_{l-1}}) \in \mathbb{R}^{N \times C}$ is obtained through a linear transformation operation $f_Q(\cdot)$, where $l$ is the layer index; (2) the key $\boldsymbol{K}^D_l = f_K(\boldsymbol{F}^P_i) \in \mathbb{R}^{\frac{H}{S_i}\frac{W}{S_i} \times C}$ and value $\boldsymbol{V}^D_l = f_V(\boldsymbol{F}^P_i) \in \mathbb{R}^{\frac{H}{S_i}\frac{W}{S_i} \times C}$ are obtained through two linear transformation operations $f_K(\cdot)$ and $f_V(\cdot)$, respectively; (3) $\boldsymbol{Q}^D_l$, $\boldsymbol{K}^D_l$, and $\boldsymbol{V}^D_l$ are subsequently processed by a masked cross-attention mechanism, expressed as follows:
\begin{equation}
\boldsymbol{X}^C_l = \mathrm{Softmax}(\boldsymbol{\boldsymbol{M}}_{l-1} + \boldsymbol{Q}_l^{D}{\boldsymbol{K}^{D}_l}^\top)\boldsymbol{V}^{D}_l + \boldsymbol{X}_{l-1},
\label{eq.transformer_decoder}
\end{equation}
where $\boldsymbol{M}_{l-1}\in\mathbb{R}^{N \times {\frac{H}{S_i}\frac{W}{S_i}}}$ (containing only the values of 0 or $-\infty$) denotes the output of the resized mask prediction obtained from the $(l-1)$-th Transformer decoder layer (readers can refer to \cite{cheng2022masked} for more details); (4) $\boldsymbol{X}_l^C$, the masked cross-attention results are subsequently processed by a self-attention mechanism, and ultimately passed through a feedforward network (FFN) to generate the query features $\boldsymbol{X}_l$. Multi-scale feature maps $\boldsymbol{F}^P_1$ to $\boldsymbol{F}^P_{n}$ are fed to their corresponding decoder layer ($n$ successive layers), and the entire process is iterated $\frac{L}{n}$ times to update the query features. Following \cite{carion2020end}, the output query features $\boldsymbol{X}_L$ produced by the Transformer decoder is mapped into a space of dimension $(K+1)$ for class predictions by a multi-layer perceptron (MLP), where $K$ represents the total number of classes to be predicted, plus an additional class representing ``no object''. The mask prediction is then obtained by performing a dot product operation between the mask embedding $\mathrm{MLP}(\boldsymbol{X}_L)$ (also generated by MLP) and the per-pixel embedding $\boldsymbol{E}$. Finally, the semantic road scene parsing result is obtained by performing a simple matrix multiplication operation (followed by an argmax function) between the mask and class predictions. Notably, this query-based decoder pipeline is identical to the approach proposed in \cite{cheng2021per}, where semantic predictions are obtained from $\boldsymbol{X}_L$.
Each query feature in $\boldsymbol{X}_L$ generates one specific mask prediction and the corresponding class predictions for $K+1$ classes, collectively forming the final semantic segmentation result.

\subsection{Loss Function}
\label{sec.loss}
We follow \cite{cheng2022masked} to train our proposed RoadFormer by minimizing the following loss function:
\begin{equation}
\mathcal{L} = \lambda_\text{mask}(\lambda_\text{ce}\mathcal{L}_\text{ce} + \lambda_\text{dice}\mathcal{L}_\text{dice}) + \lambda_\text{cls}\mathcal{L}_\text{cls},
\label{eq.loss}
\end{equation}
where $\mathcal{L}_\text{ce}$ and $\mathcal{L}_\text{dice}$ are the binary cross-entropy loss and the dice loss, respectively. $\mathcal{L}_\text{cls}$ is the classification loss, and the weighting factors $\lambda_\text{mask}$, $\lambda_\text{ce}$, $\lambda_\text{dice}$, and $\lambda_\text{cls}$ serve to balance the respective contributions of the different loss components to the overall loss. These are set in accordance with \cite{cheng2022masked}.

\section{Experiments}
\label{sec.experiments}
\begin{figure}[!t]
    \includegraphics[width=0.455\textwidth]{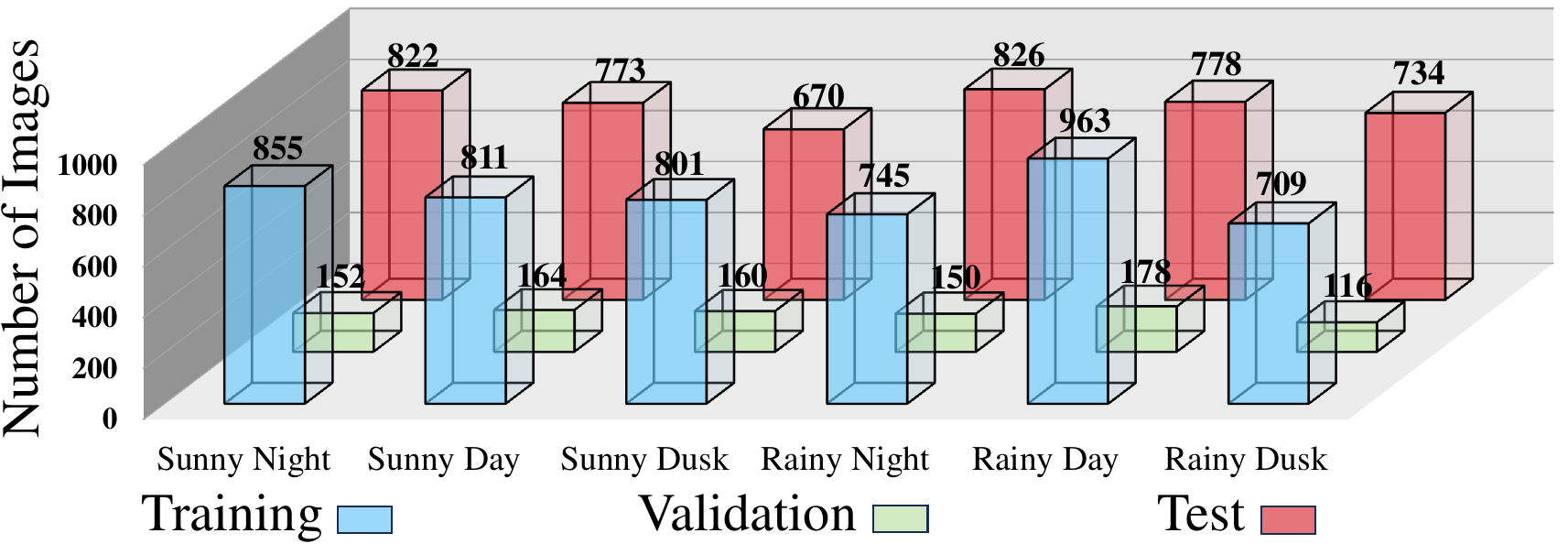}
        \caption{Data distribution in the SYN-UDTIRI dataset.}
    \label{fig:dataset_distribution}
\end{figure}

\subsection{Datasets}
\label{sec.datasets}

\subsubsection{\textbf{SYN-UDTIRI}}
\label{sec.synudtiri}
Owing to the lack of well-annotated, large-scale datasets, created specifically for road scene parsing (freespace and road defect detection), we create a synthetic dataset, referred to as SYN-UDTIRI, using the CARLA simulator \cite{dosovitskiy2017carla}. The principal contribution of this dataset lies in the integration of digital twins of road potholes acquired from the real world using our previously published 3D geometry reconstruction algorithm \cite{fan2018road, fan2021rethinking}. Moreover, to better simulate the roughness of actual roads, we introduce random Perlin noise \cite{perlin1985image} to the road data. We generate six driving scenarios, including sunny day, dusk, and night, as well as rainy day, dusk, and night, with respect to different illumination and weather conditions. Additionally, we deploy a simulated stereo rig (baseline: 0.5 m) onto a moving vehicle to acquire over 10K pairs of stereo road images (resolution: $720 \times 1,280$ pixels), along with their corresponding depth images, surface normal information, and semantic annotations, including three categories: freespace, road defect, and other objects. More details on the SYN-UDTIRI dataset are given in Fig. \ref{fig:dataset_distribution}.

\subsubsection{\textbf{KITTI Road}}
\label{sec.kitti}
The KITTI road \cite{fritsch2013new} dataset contains 289 pairs of stereo images and their corresponding LiDAR point clouds for model training and validation. It also provides a comparable amount of testing data without semantic annotations. We follow a similar data pre-processing strategy as detailed in \cite{fan2020sne}. We fine-tune our proposed RoadFormer for the test set result submission to the KITTI road benchmark. 

\subsubsection{\textbf{CityScapes}}
\label{sec.cityscapes_dataset}
The CityScapes \cite{cordts2016cityscapes} dataset is a widely utilized urban scene dataset, containing 2,975 stereo training images and 500 validation images, with well-annotated semantic annotations. Due to the limited sample size of the KITTI road dataset, we conduct additional experiments on the CityScapes dataset to further demonstrate the effectiveness of our proposed RoadFormer on large-scale datasets. All experimental results are obtained using the validation set since ground-truth annotations are not provided on the test set. Quantitative and qualitative evaluation results on the test set are typically acquired by submitting results to the online CityScapes benchmark suite. Furthermore, given our specific focus on road scene parsing, we have to reorganize the dataset to train and evaluate models for only two classes: road and others. It is noteworthy that the corresponding surface normal information is derived from depth images obtained using RAFT-Stereo \cite{lipson2021raft} trained on the KITTI \cite{menze2015object} dataset. 

\subsubsection{\textbf{ORFD}}
\label{sec.orfd}
The ORFD \cite{min2022orfd} dataset is designed specifically for off-road freespace detection. It contains 12,198 RGB images and their corresponding LiDAR point clouds, collected across various scenes, under different weather and illumination conditions. We follow the data splitting and pre-processing strategies (except for surface normal estimation) detailed in \cite{min2022orfd} for our experiments.

\subsection{Experimental Setup and Evaluation Metrics} 
\label{sec.experimental_setup_evaluation_metrics}

In our experiments, we compare our proposed RoadFormer with four single-modal networks and five data-fusion networks. The single-modal networks are trained using RGB images, while the data-fusion networks are trained using both RGB images and surface normal information (estimated using D2NT \cite{icra_2023_d2nt} owing to its superior accuracy compared to other methods). All the networks undergo training for the same number of epochs. For RoadFormer training, we utilize the AdamW optimizer \cite{loshchilov2018decoupled} with a polynomial learning rate decay strategy \cite{chen2017deeplab}. The learning rate begins at $10^{-4}$ with a weight decay of $5 \times 10^{-2}$. We apply learning rate multipliers of $10^{-1}$ to both ConvNeXt \cite{liu2022convnet} and Swin \cite{liu2021swin} backbones.

\begin{figure*}[t!]
    \includegraphics[width=0.98\textwidth]{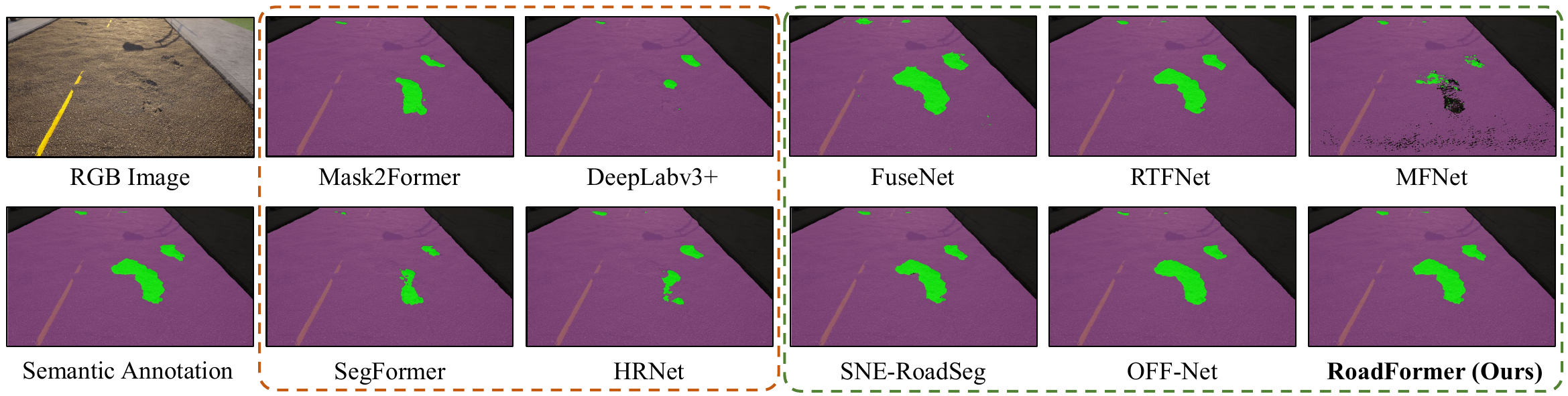}
        \caption{Qualitative comparison between RoadFormer and other SoTA networks on the SYN-UDTIRI dataset. The freespace, road defect, and background areas are shown in purple, green, and black, respectively.}
    \label{fig.carla_viz}
\end{figure*}

We employ five common metrics to quantify the network performance: accuracy (Acc), precision (Pre), recall (Rec), intersection over union (IoU), and F-score (Fsc). We refer readers to our previous work \cite{fan2020sne} for more details on these metrics. Additionally, the evaluation metrics used for the KITTI road benchmark can be found on its official webpage: \url{https://www.cvlibs.net/datasets/kitti/eval_road.php}.

\begin{table}[htbp]
\settablefont
\centering
\caption{Ablation study on backbone selection. 
}
\renewcommand{\arraystretch}{1.25}
{
\setlength{\tabcolsep}{8.3pt}
\begin{tabular}{C{1.2cm}|C{0.9cm}|cccc}
\toprule[1pt]
Dataset & Backbone & IoU (\%)& Fsc (\%)& Pre (\%)& Rec (\%)\\
\hline
\hline
\multirow{2}{*}{\rotatebox[origin=c]{0}{SYN-UDTIRI}}
&ConvNeXt & \textbf{93.38} & \textbf{96.58} & 96.58 & \textbf{96.74}  \\
&Swin & 93.18 & 96.47 & \textbf{96.59} & 96.35 \\
\hline
\multirow{2}{*}{\rotatebox[origin=c]{0}{CityScapes}}
&ConvNeXt & \textbf{95.80} & \textbf{97.86} & \textbf{97.74} & \textbf{97.97} \\
&Swin & 94.69 & 97.27 & 97.25 & 97.29 \\
\bottomrule[1pt]
\end{tabular}
}
\label{tab.backbone_ablation}
\end{table}
\begin{table}[htbp]
\settablefont
\centering
\caption{Ablation study demonstrating the effectiveness of our proposed HFFM and FFRM. 
}
\setlength{\tabcolsep}{4pt}
{
\begin{tabular}{C{0.45cm}|C{0.6cm}|C{0.6cm}|ccccc|c}
\toprule[1pt]
SEB & HFFM & FFRM & IoU (\%) & Acc (\%) & Fsc (\%) & Pre (\%)& Rec (\%) & FPS \\
\hline
\hline
$\times$ &$\times$ & $\times$ & 95.11 & 96.87 & 97.50 & 98.12 & 96.87 & 21.80 \\
$\checkmark$ & $\times$ & $\times$ & 95.45 & 97.40 & 97.67 & \textbf{97.95} & 97.40 & 21.60 \\
$\times$  &$\times$ & $\checkmark$ & 95.49 & 97.59 & 97.69 & 97.79 & 97.59 & 21.60 \\
$\times$ &$\checkmark$ & $\times$ & 95.34 & 97.67 & 97.61 & 97.55 & 97.67 & 20.50 \\
$\times$ &$\checkmark$ & $\checkmark$ & \textbf{95.80} & \textbf{97.97} & \textbf{97.86} & 97.74 & \textbf{97.97} & 20.10 \\
\bottomrule[1pt]
\end{tabular}
}
\label{tab.modules_ablation}
\end{table}

\begin{figure*}[!t]
    \includegraphics[width=0.98\textwidth]{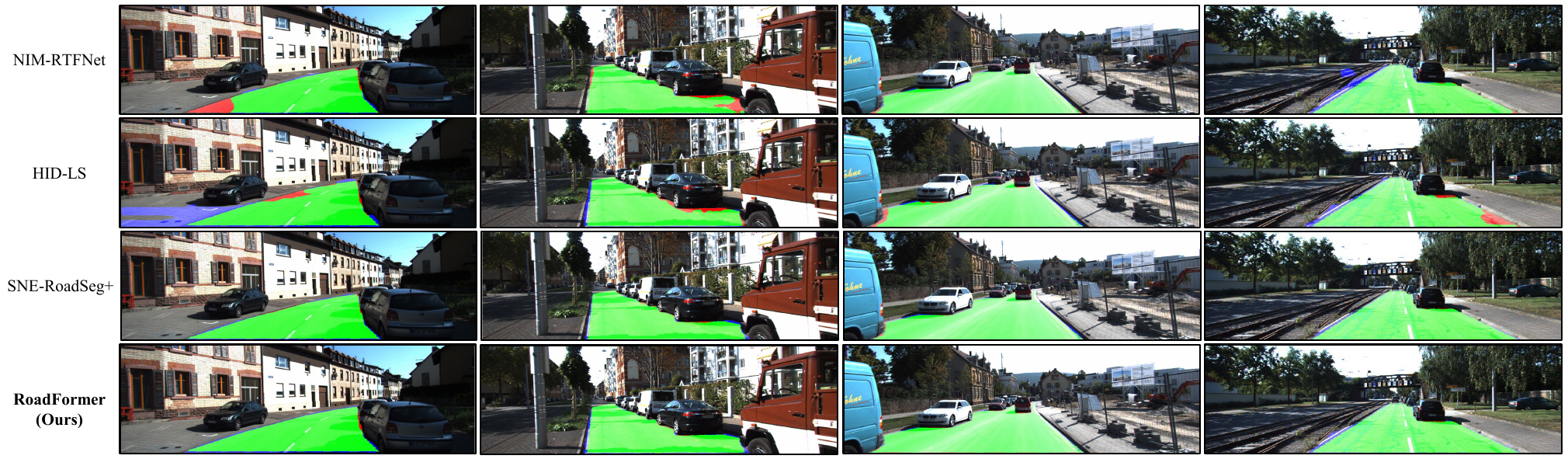}
        \caption{Qualitative comparison between RoadFormer and other SoTA networks on the KITTI road test set. The results are obtained from the official KITTI online benchmark suite. True-positive, false-positive, and false-negative classifications are shown in green, blue, and red, respectively.}
    \label{fig.kitti_viz}
\end{figure*}

\begin{figure*}[t!]
    \includegraphics[width=0.98\textwidth]{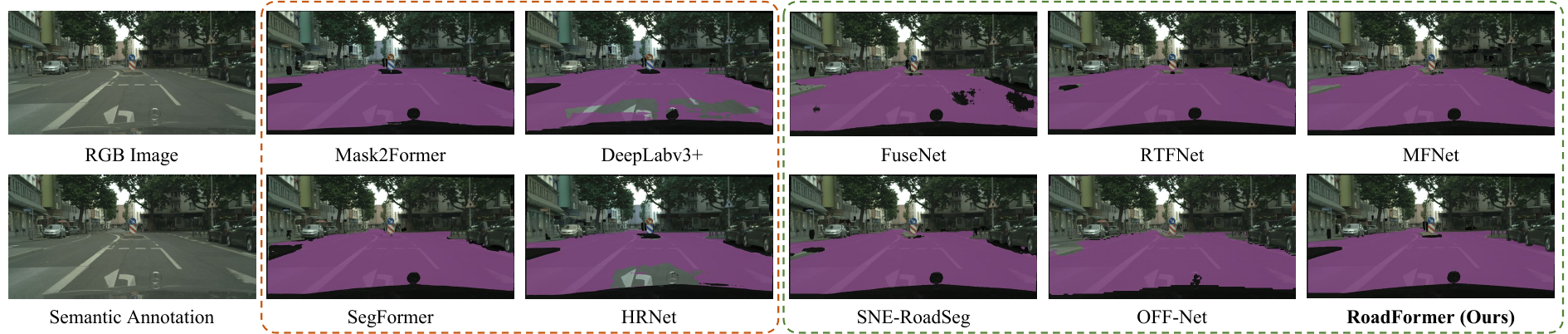}
        \caption{Qualitative comparison between RoadFormer and other SoTA networks on the CityScapes dataset. The freespace and ignored areas are shown in purple and black, respectively.}
    \label{fig.cityscapes_viz}
\end{figure*}

\begin{figure*}[t!]
    \includegraphics[width=0.98\textwidth]{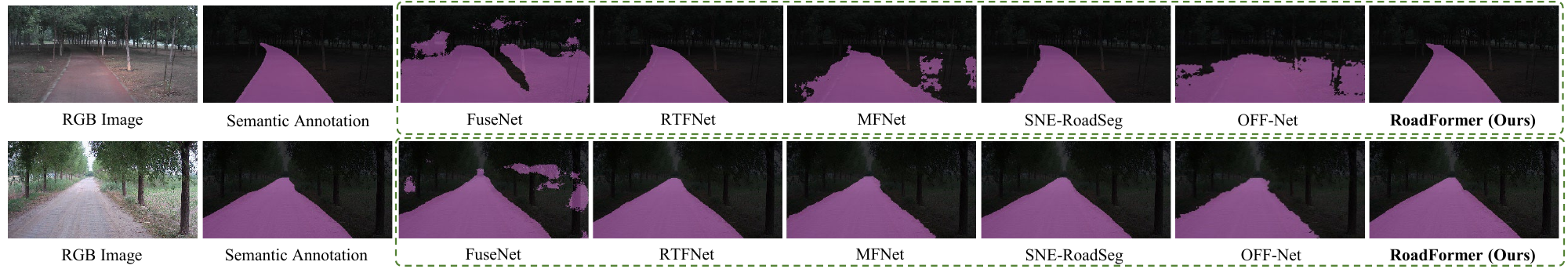}
        \caption{Qualitative comparison between RoadFormer and other SoTA data-fusion networks on the ORFD dataset. The freespace and remaining areas are shown in purple and black, respectively.}
    \label{fig.orfd_viz}
\end{figure*}

\subsection{Ablation Study}
\label{sec.ablation_study}
Although recent studies on scene parsing, such as \cite{cheng2022masked} and \cite{jain2023oneformer}, often combine the widely used Swin Transformer encoder with a Transformer decoder, we hypothesize that the Swin Transformer might not be the optimal choice for all parsing tasks. Therefore, we perform an ablation study on SYN-UDTIRI and CityScapes to compare the performance of ConvNeXt and Swin Transformer for encoder backbone selection. As shown in Table \ref{tab.backbone_ablation}, ConvNeXt demonstrates superior performance over Swin, achieving improvements of $0.20\%$-$1.11\%$ in terms of IoU and $0.11\%$-$0.59\%$ in terms of F-score. These results suggest that our proposed RoadFormer is compatible with both CNN-based and Transformer-based backbones, and ConvNeXt is generally a preferable option for road scene parsing. Therefore, in the following experiments, we will utilize ConvNeXt as our backbone network. 

Given that the design of FFRM is inspired by SEB, we first compare the performance of our RoadFormer when incorporated with either SEB or FFRM. Furthermore, we also evaluate the individual effectiveness of HFFM and FFRM, as well as their compatibility. The results shown in Table \ref{tab.modules_ablation} indicate that (1) FFRM outperforms SEB, achieving an improvement of $0.04\%$ in terms of IoU, (2) HFFM yields a $0.23\%$ higher IoU compared to the baseline setup, and (3) the combined utilization of HFFM and FFRM modules results in better performance than using these modules independently.

Despite our HFSB introducing a slight overhead on inference speed due to its multi-scale recalibration and fusion of heterogeneous features, RoadFormer achieves an inference speed of $\sim20$ FPS when processing images at a resolution of $352 \times 640$ pixels on an NVIDIA RTX 3090 GPU. This performance effectively meets real-time processing requirements.

\subsection{Experimental Results}
\label{sec.experiment_detail}

\begin{table}[htbp]

\settablefont
\centering
\caption{Comparison of SoTA algorithms published on the KITTI road benchmark.}

{
\setlength{\tabcolsep}{7.2pt}
\begin{tabular}{l|c|c|c|c|c}
\toprule[1pt]
Method & MaxF (\%) & AP (\%) & Pre (\%) & Rec (\%) & Rank \\ 
\hline
\hline
NIM-RTFNet \cite{wang2020applying} & 96.02 & 94.01 & 96.43 & 95.62 & 13 \\ 
HID-LS \cite{gu2018histograms} & 93.11 & 87.33 & 92.52 & 93.71 & 33 \\ 
LC-CRF \cite{gu2019road} & 95.68 & 88.34 & 93.62 & 97.83 & 15 \\ 
SNE-RoadSeg \cite{fan2020sne} & 96.75 & 94.07 & 96.90 & 96.61 & 8 \\ 
SNE-RoadSeg+ \cite{wang2021sne} & 97.50 & 93.98 & \textbf{97.41} & 97.58 & 2 \\
PLB-RD \cite{sun2022pseudo} & 97.42 & \textbf{94.09} & 97.30 & 97.54 & 3 \\
LRDNet+ \cite{khan2022lrdnet} & 96.95 & 92.22 & 96.88 & 97.02 & 4 \\
DFM-RTFNet \cite{wang2021dynamic} & 96.78 & 94.05 & 96.62 & 96.93 & 7 \\
\hline
\textbf{RoadFormer} & \textbf{97.50} & 93.85 & 97.16 & \textbf{97.84} & {\color{red}\textbf{1}} \\
\bottomrule[1pt]
\end{tabular}
}

\label{table.kitti}
\end{table}

\begin{table}[htbp]

\settablefont
\centering
\caption{Quantitative results on the SYN-UDTIRI dataset. 
}
\renewcommand{\arraystretch}{1.05}
{
\begin{tabular}{
l|c|l|cccc
}
\toprule[1pt]
& Subset & Method & IoU (\%)& Fsc (\%)& Pre (\%)& Rec (\%)\\
\hline
\hline
\multirow{8}{*}{\rotatebox[origin=c]{90}{RGB}}
& \multirow{4}{*}{\rotatebox[origin=c]{90}{Validation}} & Mask2Former& 64.29 & 78.27  & 83.0 & 74.05 \\
& & SegFormer & 52.46 & 68.82 & 70.13 & 67.55 \\
& & DeepLabv3+ & 52.94 & 69.23 & 75.23 & 64.12 \\
& & HRNet & 52.92 & 69.21 & 79.46 & 61.30 \\
\cline{2-7}
& \multirow{4}{*}{\rotatebox[origin=c]{90}{Test}} & Mask2Former & 46.91 & 63.87 & 73.59 & 56.41 \\
& & SegFormer & 36.34 & 53.31 & 57.23 & 49.89 \\
& & DeepLabv3+ & 34.76 & 51.58 & 62.54 & 43.90 \\
& & HRNet & 35.47 & 52.37 & 69.09 & 42.16 \\
\hline
\multirow{12}{*}{\rotatebox[origin=c]{90}{RGB-Normal}} 
& \multirow{6}{*}{\rotatebox[origin=c]{90}{Validation}} & FuseNet & 67.30 & 80.40 & 68.30  & \textbf{97.80} \\
& & SNE-RoadSeg & 92.00 & 95.80 & 96.30  & 95.40 \\
& & RTFNet & 90.30 & 94.90 & 94.10 & 95.70 \\
& & OFF-Net & 83.90 & 91.30 & 91.70 & 90.80 \\
& & MFNet & 89.50 & 94.50 & 95.70 & 93.30 \\
\cline{3-7}
& & \textbf{RoadFormer} & \textbf{93.35} & \textbf{96.56} & \textbf{96.53} & 96.59 \\
\cline{2-7}
& \multirow{6}{*}{\rotatebox[origin=c]{90}{Test}} & FuseNet & 70.70 & 82.90 & 72.10  & \textbf{97.50} \\
& & SNE-RoadSeg & 92.10 & 95.90 & \textbf{96.70}  & 95.10 \\
& & RTFNet & 90.50 & 95.00 & 95.50 & 94.50 \\
& & OFF-Net & 83.80 & 91.20 & 91.90 & 90.50 \\
& & MFNet & 87.70 & 93.50 & 96.20 & 90.90 \\
\cline{3-7}
& & \textbf{RoadFormer} & \textbf{93.51} & \textbf{96.65} & 96.61 & 96.69 \\
\bottomrule[1pt]
\end{tabular}
}
\label{table.synudtiri}
\end{table}

The quantitative results on the SYN-UDTIRI, CityScapes, ORFD, and KITTI Road datasets are presented in Tables \ref{table.kitti}-\ref{table.orfd}. Additionally, we also present readers with the qualitative results on these four datasets in Figs. \ref{fig.carla_viz}-\ref{fig.orfd_viz}. These results suggest that our proposed RoadFormer outperforms all other SoTA networks across all four datasets, demonstrating its exceptional performance and robustness in effectively parsing various types of road scenes, including synthetic roads with defects, urban roads, and rural roads.

As shown in Table \ref{table.synudtiri}, it is evident that data-fusion networks demonstrate considerably improved robustness compared to single-modal networks, particularly in road defect detection. We exclude the freespace detection results for comparison due to the closely matched performance across all networks in this task (the single-modal and data-fusion networks achieve IoUs of over 98.5\% and 99.5\%, respectively). As expected, data-fusion networks achieve better generalizability compared to single-modal networks. This improvement is attributed to geometric features extracted from surface normal information. 

\begin{table}[t!]
\settablefont
\centering
\caption{Quantitative results on the CityScapes dataset.}
\renewcommand{\arraystretch}{1.05}
{
\setlength{\tabcolsep}{5pt}
\begin{tabular}{
l|L{1.8cm}|C{0.76cm}C{0.76cm}C{0.76cm}C{0.76cm}|C{0.95cm}
}
\toprule[1pt]
& Method  & IoU (\%) & Fsc (\%) & Pre  (\%) & Rec (\%) & mIoU (\%) \\
\hline
\hline
\multirow{4}{*}{\rotatebox[origin=c]{90}{RGB}}
& Mask2Former & 93.84 & 96.82 & 97.14  & 96.51 & 74.80 \\
& SegFormer & 93.98 & 96.90 & 96.02 & 97.79 & 64.51 \\
& DeepLabv3+ & 93.82 & 96.81 & 96.99  & 96.63 & 68.66 \\
& HRNet & 94.06 & 96.94 & 96.29 & 97.59 & 70.10 \\
\hline
\multirow{6}{*}{\rotatebox[origin=c]{90}{RGB-Normal}} 
& FuseNet & 91.60 & 95.60 & 96.00 & 95.30 & 52.70 \\
& SNE-RoadSeg & 93.80 & 96.80 & 96.10 & 97.50 & 53.40 \\
& RTFNet & 94.10 & 96.90 & 96.30 & 97.60 & 49.60 \\
& OFF-Net & 89.60 & 94.50 & 93.40 & 95.70 & 39.20 \\
& MFNet & 92.10 & 95.90 & 94.10 & 97.70 & 49.30 \\
\cline{2-7}
& \textbf{RoadFormer} & \textbf{95.80} & \textbf{97.86} & \textbf{97.74} & \textbf{97.97} & \textbf{76.20} \\
\bottomrule[1pt]
\end{tabular}

}
\label{table.cityscapes}
\end{table}

Additionally, the results on the CityScapes dataset somewhat exceed our expectations. The performance of the single-modal networks, with the IoU fluctuating within a range of $0.2\%$, is quite comparable to that of the data-fusion networks. To our surprise, all single-modal networks outperform data-fusion networks, except for RTFNet and our RoadFormer. We speculate that these unexpected results could be attributed to inaccuracies in the disparity maps used for surface normal estimation, as they are directly obtained from pre-trained stereo matching networks. The previous data-fusion networks employ basic element-wise addition or feature concatenation operations for feature fusion, leading to performance degradation when surface normal information is inaccurate. In contrast, benefiting from the adaptive fusion and recalibration of heterogeneous features via our designed HFSB, our RoadFormer achieves the highest scores across all metrics, including the mean IoU (mIoU) computed across $20$ categories (including the ``ignore" category) in the full-pixel semantic segmentation task.

\begin{table}[htbp]
\settablefont
\centering
\caption{Quantitative results on the test set of ORFD dataset.
We present both the results reported in the original paper \cite{min2022orfd} and those obtained through our re-implementation. $\dagger$ denotes the model trained using RGB-Depth data in the original implementation.
}
{
\setlength{\tabcolsep}{9.5pt}
\begin{tabular}{
l|l|ccccc
}
\toprule[1pt]
& Method  & IoU (\%)& Fsc (\%) & Pre  (\%)& Rec (\%)\\
\hline
\hline
\multirow{3}{*}{\rotatebox[origin=c]{90}{Published}}
& FuseNet$^\dagger$ & 66.00  & 79.50 & 74.50 & 85.20 \\
& SNE-RoadSeg & 81.20  & 89.60  & 86.70  & 92.70  \\
& OFF-Net & 82.30  & 90.30  & 86.60 & 94.30 \\
\hline
\multirow{6}{*}{\rotatebox[origin=c]{90}{Re-implemented}} 
& FuseNet & 59.00 & 74.20 & 59.30 & \textbf{99.10} \\
& SNE-RoadSeg & 79.50 & 88.60 & 90.30 & 86.90 \\
& RTFNet  & 90.70 & 95.10 & 93.80 & 96.50 \\
& OFF-Net & 81.80 & 90.00 & 84.20 & 96.70 \\
& MFNet & 81.70 & 89.90 & 89.60 & 90.30 \\
\cline{2-6}
& \textbf{RoadFormer} & \textbf{92.51} & \textbf{96.11} & \textbf{95.08} & 97.17 \\
\bottomrule[1pt]
\end{tabular}%
}
\label{table.orfd}
\end{table}

To ensure a fair comparison with SoTA networks on the ORFD dataset, we present both the results reported in the original paper \cite{min2022orfd} for three networks and those obtained through our re-implementation for six networks, as shown in Table \ref{table.orfd}. We observe that the performances of data-fusion networks differ significantly on the ORFD dataset. This is likely due to the challenge of labeling accurate off-road semantic ground truth. The inaccurate annotations may introduce ambiguities during the model training process. Finally, we submit the test set results produced by RoadFormer to the KITTI road online benchmark for performance comparison. As shown in Table \ref{table.kitti}, RoadFormer demonstrates superior performance compared to all previously published methods.

\section{Conclusion and Future Work}
\label{sec.conclusion}

This article presented RoadFormer, a powerful data-fusion Transformer architecture designed for road scene parsing. It contains a duplex encoder, a novel feature synergy block, and Transformer-based decoders. Compared to previous works, RoadFormer demonstrates the effective fusion of heterogeneous features and improved accuracy. It outperforms all existing semantic segmentation networks on our newly created SYN-UDTIRI dataset and three public datasets, while ranking first on the KITTI road benchmark upon submission. As critical components of RoadFormer, feature fusion using self-attention has proven superior to pure CNNs for road scene parsing, and we aim to investigate its potential for more common scene parsing tasks in the future. On the other hand, although achieving high accuracy, the real-time performance of data-fusion networks still needs improvement, which we will leave to future work.

\normalem
\bibliographystyle{IEEEtran}
\bibliography{refs}

\end{document}